# Price Prediction in a Trading Agent Competition

**Michael P. Wellman**                                    WELLMAN@UMICH.EDU
**Daniel M. Reeves**                                      DREEVES@UMICH.EDU
**Kevin M. Lochner**                                      KLOCHNER@UMICH.EDU
**Yevgeniy Vorobeychik**                                  YVOROBEY@UMICH.EDU
*University of Michigan, Artificial Intelligence Laboratory*
*Ann Arbor, MI  48109-2110  USA*

## Abstract

The 2002 Trading Agent Competition (TAC) presented a challenging market game in the domain of travel shopping. One of the pivotal issues in this domain is uncertainty about hotel prices, which have a significant influence on the relative cost of alternative trip schedules. Thus, virtually all participants employ some method for predicting hotel prices. We survey approaches employed in the tournament, finding that agents apply an interesting diversity of techniques, taking into account differing sources of evidence bearing on prices. Based on data provided by entrants on their agents' actual predictions in the TAC-02 finals and semifinals, we analyze the relative efficacy of these approaches. The results show that taking into account game-specific information about flight prices is a major distinguishing factor. Machine learning methods effectively induce the relationship between flight and hotel prices from game data, and a purely analytical approach based on competitive equilibrium analysis achieves equal accuracy with no historical data. Employing a new measure of prediction quality, we relate absolute accuracy to bottom-line performance in the game.

## 1. Introduction

Many market decision problems involve some anticipation or forecast of future prices. Price prediction is particularly important, for example, in committing to a binding offer to purchase a good that has complements to be purchased at a later date. This sort of scenario arises whenever there are sequential or overlapping auctions for related goods. Although market forecasting techniques are in widespread use over a broad range of applications, we are unaware of studies exploring the problem in a context reminiscent of multi-auction environments.

The annual Trading Agent Competition (TAC) (Wellman et al., 2003) provides a convenient medium for studying approaches to price prediction. As an open-invitation tournament, it has attracted a diverse community of researchers interested in many aspects of trading agent strategy. Price prediction has turned out to be a pivotal issue in the TAC market game,[1] and an interesting array of approaches has emerged from agent designers' efforts over three years of competition. Since TAC defines a controlled, regular, repeatable, and transparent environment for observing trading agent behavior, it is also uncommonly amenable to analysis.

---

1. We refer to the original ("classic") TAC game, a scenario in the domain of travel shopping. Sadeh et al. (2003) introduced a second game ("TAC/SCM"), in the domain of supply chain management, and we expect that further domains will be the subject of trading competitions in coming years.

## 2. TAC Travel-Shopping Game

The TAC market game presents a travel-shopping task, where traders assemble flights, hotels, and entertainment into trips for a set of eight probabilistically generated clients. Clients are described by their preferred arrival and departure days ($pa$ and $pd$), the premium ($hp$) they are willing to pay to stay at the "Towers" (T) hotel rather than "Shanties" (S), and their values for three different types of entertainment events. The agents' objective is to maximize the value of trips for their clients, net of expenditures in the markets for travel goods.

*Flights*. A feasible trip includes round-trip air, which consists of an inflight day $i$ and outflight day $j$, $1 \leq i < j \leq 5$. Flights in and out each day are sold independently, at prices determined by a stochastic process. The initial price for each flight is $\sim U[250, 400]$, and follows a random walk thereafter with an increasingly upward bias.

*Hotels*. Feasible trips must also include a room in one of the two hotels for each night of the client's stay. There are 16 rooms available in each hotel each night, and these are sold through ascending 16th-price auctions. When the auction closes, the units are allocated to the 16 highest offers, with all bidders paying the price of the lowest winning offer. Each minute, the hotel auctions issue *quotes*, indicating the 16th- (*ASK*) and 17th-highest (*BID*) prices among the currently active unit offers. Each minute, starting at 4:00, one of the hotel auctions is selected at random to close, with the others remaining active and open for bids.

*Entertainment*. Agents receive an initial random allocation of entertainment tickets (indexed by type and day), which they may allocate to their own clients or sell to other agents through continuous double auctions.

A feasible client trip $r$ is defined by an inflight day $in_r$, outflight day $out_r$, and hotel type ($H_r$, which is 1 if T and 0 if S). Trips also specify entertainment allocations, but for purposes of this paper we summarize expected entertainment surplus $\phi$ as a function of trip days. (The paper describing our agent (Cheng et al., 2004) provides details on $\phi(r)$ as well as some other constructs glossed here.) The value of this trip for client $c$ (with preferences $pa$, $pd$, $hp$) is then given by

$$v_c(r) = 1000 - 100(|pa - in_r| + |pd - out_r|) + hp \cdot H_r + \phi(r). \tag{1}$$

At the end of a game instance, the TAC server calculates the optimal allocation of trips to clients for each agent, given final holdings of flights, hotels, and entertainment. The agent's game score is its total client trip utility, minus net expenditures in the TAC auctions.

## 3. Price Prediction

TAC participants recognized early on the importance of accurate price prediction in overall performance (Stone & Greenwald, 2004). The prices of hotels are highly variable from game to game, yet a hotel's price is not finalized until its auction closes—some minutes into the game, depending on the random closing order. Because agents tend not to submit serious hotel bids until the first closing is imminent, no useful information is revealed by price quotes until the fourth minute of play. Complementarity among goods dictates that outcomes of early auctions significantly affect the value an agent places on a particular hotel later in the game, and conversely, the prices of hotels available later dictate whether an agent had bid wisely early in the game.

Anticipating hotel prices is a key element in several decisions facing a TAC agent, in particular:

1. *Selecting trip itineraries.* Because flight prices tend to increase, agents have a great incentive to commit to traveling on particular days early in the game. Yet the quality of day choices depends crucially on the hotel prices of the included travel days.

2. *Bidding policy.* The likelihood of obtaining a good with a given bid depends obviously on the resulting clearing price. Moreover, the value of any particular good is in general a function of the price of others. For example, the value of obtaining a room at hotel (S,$i$) is an increasing function of the projected cost of the alternative hotel on that day, (T,$i$), and a decreasing function of the projected cost of complementary hotel rooms on the adjacent days, (S,$i - 1$) and (S,$i + 1$).

Given the importance of price prediction, it is not surprising that TAC researchers have explored a variety of approaches. Stone et al. (2003) noted a diversity of price estimation methods among TAC-01 agents. Our particular interest in the problem is certainly connected to our own TAC-02 agent, Walverine, which introduced a method quite distinct from those reported previously. Thus, we were highly motivated to characterize performance on this price prediction task.

Indeed, if competitions such as TAC are to be successful in facilitating research, it will be necessary to separately evaluate techniques developed for problem subtasks (Stone, 2002). Although interesting subtasks tend not to be strictly separable in such a complex game, the price-prediction component of trading strategy may be easier to isolate than most. In particular, the problem can be formulated in its own terms, with natural absolute accuracy measures. And in fact, as we see below, most agent developers independently chose to define price prediction as a distinct task in their own agent designs.

We divide price prediction into two phases: *initial* and *interim*. Initial refers to the beginning of the game, before any hotel auctions close or provide quote information. Interim refers to the method employed thereafter. Since the information available for initial prediction (flight prices, client preferences—see Section 5.2) is a strict subset of that available for interim (which adds transaction and hotel price data), most agents treat initial prediction as just a (simpler) special case.

Initial prediction is relevant to bidding policy for the first hotel closing, and especially salient for trip choices as these are typically made early in the game. Interim prediction supports ongoing revision of bids as the hotel auctions start to close. Our analysis and report focus on initial prediction, mainly because it is the simpler of the two tasks, involving less potential information. Moreover, agents initially have relatively comparable information sets, thus providing for a cleaner analysis. Interim prediction is also quite important and interesting, and should be the focus of further work.

## 4. TAC-02 Agents

The nineteen agents who participated in the TAC-02 tournament are listed in Table 1. The table presents raw average scores from the finals and semifinals, and weighted averages for the seeding round. For overviews of most of these agents, see the survey edited by Greenwald (2003a).

The seeding rounds were held during the period 1-12 July, each agent playing 440 games. Weights increased each day, so that later games counted more than earlier, and the lowest 10 scores for each agent were dropped. The top 16 agents advanced to the semifinals, held on 28 July in Edmonton, Canada. There were two semifinal heats: H1 comprising agents seeded 1-4 and 13-16, with the 5-12 seeds placed in heat H2. The top four teams from each heat (14 games, lowest score dropped) proceeded to the finals, which ran for 32 games the same day. Further details about the TAC-02 tournament are available at `http://www.sics.se/tac`.

| Agent | Affiliation | Seeding | Semifinals | Finals |
|-------|-------------|---------|-----------|--------|
| ATTac | AT&T Research (et al.) | 3131 | H1: 3137 | — |
| BigRed | McGill U | 696 | — | — |
| cuhk | Chinese U Hong Kong | 3055 | H2: 3266 | 3069 |
| harami | Bogazici U | 2064 | — | — |
| kavayaH | Oracle India | 2549 | H1: 3200 | 3099 |
| livingagents | Living Systems AG | 3091 | H1: 3310 | 3181 |
| PackaTAC | N Carolina State U | 2835 | H2: 3250 | — |
| PainInNEC | NEC Research (et al.) | 2319 | H1: 2193 | — |
| RoxyBot | Brown U | 2855 | H2: 3160 | — |
| sics | Swedish Inst Comp Sci | 2847 | H2: 3146 | — |
| SouthamptonTAC | U Southampton | 3129 | H1: 3397 | 3385 |
| Thalis | U Essex | 3000 | H2: 3199 | 3246 |
| tniTac | Poli Bucharest | 2232 | H1: 3108 | — |
| TOMAhack | U Toronto | 2809 | H2: 2843 | — |
| tvad | Technion | 2618 | H1: 2724 | — |
| UMBCTAC | U Maryland Baltimore Cty | 3118 | H1: 3208 | 3236 |
| Walverine | U Michigan | 2772 | H2: 3287 | 3210 |
| WhiteBear | Cornell U | 2966 | H2: 3324 | 3413 |
| zepp | Poli Bucharest | 2098 | — | — |

Table 1: TAC-02 tournament participants, and their scores by round.

## 5. Price Prediction Survey

Shortly after the TAC-02 event, we distributed a survey to all the entrants eliciting descriptions and data documenting their agents' price-prediction methods. Sixteen out of 19 teams responded to the survey, including 14 of 16 semifinalists, and all eight finalists. The result provides a detailed picture of the prediction techniques employed, and enables some comparison of their efficacy with respect to a common experience—the TAC-02 finals and semifinals.

Thirteen out of the 16 respondents reported that their agents did indeed form explicit price predictions for use in their trading strategies. These thirteen are listed in Table 2, along with some high-level descriptors of their approach to the initial prediction task.[2] In addition, tniTac and zepp responded that price predictions were part of their agent designs, but were not developed sufficiently to be deployed in the tournament. TOMAhack reported an ambitious design (also not actually employed) based on model-free policy learning, which does account for other agents' bidding behavior but without formulating explicit price predictions.

---

2. Though we address only initial prediction in this report, our survey also solicited descriptions about interim methods.

| Agent | Approach | Form | Notes |
|---|---|---|---|
| ATTac | machine learning | prob | boosting |
| cuhk | historical | priceline | moving average |
| harami | historical | prob | |
| kavayaH | machine learning | point | neural net |
| livingagents | historical | point | |
| PackaTAC | historical | prob | |
| RoxyBot | historical | prob | |
| sics | historical | priceline | |
| SouthamptonTAC | historical | point | classification to reference categories |
| Thalis | historical (?) | point | survey incomplete |
| UMBCTAC | historical | point | |
| Walverine | competitive | point | equilibrium analysis |
| WhiteBear | historical | point | |

Table 2: Agents reporting prediction of hotel prices in TAC-02.

## 5.1 Forms of Prediction

One distinction observed in TAC-01 was that some agents explicitly formulated predictions in terms of probability distributions over prices, rather than point estimates. Predictions in this form enable the agent to properly account for price uncertainty in decision making. Thus, we asked entrants about the form of their predictions in our survey.

Although most agents generate point predictions, there are notable exceptions. ATTac's boosting algorithm (Stone et al., 2003) expressly learns probability distributions associated with game features. RoxyBot tabulates game price statistics for a direct estimation of deciles for each hotel auction. PackaTAC and harami measure historical variance, combining this with historical averaging to define a parametric distribution for each hotel price. Walverine predicts point prices, but its "hedging" approach for some decisions amounts to forming an effective distribution around them.

Given a prediction in the form of a distribution, agents may make decisions by sampling or through other decision-analytic techniques. The distribution may also facilitate the interim prediction task, enabling updates based on treating observations such as price quotes as evidence. However, the first controlled experiment evaluating the distribution feature, in the context of ATTac (Stone et al., 2003), did not find an overall advantage to decision-making based on distributions compared to using mean values. The authors offered several possible explanations for the observed performance, including (1) that their implementation employs insufficient samples, and (2) that their use of distributions makes the unrealistic assumption that subsequent decisions can be made with full knowledge of the actual price values. Greenwald (2003b) also found that bidding marginal utility based on means outperformed bidding expected marginal utility based on distributions, this time implemented in the context of RoxyBot. We have since performed analogous trials using Walverine—which generates and applies distributions in yet a third way—and also found bidding based on means to be superior to the distribution-based bidding the agent actually employed in TAC-02. Although the source of this deficiency has not been conclusively established, we speculate that the second reason adduced by the ATTac designers is most plausible.

Despite this evidence, alternative ways of using the distributions may well prove beneficial. The study by Greenwald (2003b) demonstrated an advantage to RoxyBot's 2002 strategy of evaluating candidate bid sets with respect to distributions, compared to its 2000 strategy of evaluating them with respect to means.

Nevertheless, for agents that predict probability distributions, we take the mean of these distributions as the subject of our analysis. This may discount potential advantages, but based on the discussion above, we suspect that—with the possible exception of RoxyBot—agents did not actually benefit from predicting distributions in TAC-02.

Another variation in form is the prediction of prices as a function of quantity demanded. From the first TAC, entrants recognized that purchasing additional units may cause the price to increase, and so introduced the concept of *pricelines*, which express estimated prices by unit (Boyan & Greenwald, 2001; Stone & Greenwald, 2004). Agents sics and cuhk reported predicting pricelines.[3] In both cases, the agent started with a baseline point prediction for the first unit of each hotel, and derived the remainder of the priceline according to some rule. For example, sics predicted the price for the $n$th unit (i.e., price given it demands $n$ units) to be $px^{n-1}$, where $p$ is the baseline prediction and $x$ is 1.15 for hotels on day 1 or 4, and 1.25 for hotels on day 2 or 3.

In the succeeding analysis, we evaluate predictions in terms of baseline prices only. As noted below, our accuracy measures applied to pricelines would not reflect their actual value.

## 5.2 Information Employed

The set of information potentially available at the beginning of the game includes all data from past games, the initial vector of flight prices, and the agent's own client preferences. For TAC-02, all agents except Walverine reported using historical information in their predictions. Only ATTac, kavayaH, and Walverine employ flight prices. All agents that construct pricelines effectively take account of own client preferences. Walverine does not construct pricelines but does factor in its own client preferences as part of its equilibrium calculations.

The identities of other agents participating in a game instance are not known during the TAC preliminary (qualifying and seeding) rounds, as agents are drawn randomly into a round-robin tournament. However, the semifinal and final rounds fixed a set of eight agents for a series of games, and so the identity of other agents was effectively observable. ATTac is the only agent to exploit the availability of this information.

## 6. Approaches to Price Prediction

Based on survey responses, we divide TAC-02 prediction techniques into three categories.

### 6.1 Historical Averaging

Most agents took a relatively straightforward approach to initial price prediction, estimating the hotel clearing prices according to observed historical averages. For example, harami calculates the mean hotel prices for the preceding 200 games, and uses this as its initial prediction. The respective agents classified as adopting the "historical" approach in Table 2 differ on what set of games they include in the average, but most used games from the seeding round. Given a dataset, agents tend to use the sample mean or distribution itself as the estimate,[4] at least as the baseline.

---

3. WhiteBear also reported using pricelines for interim prediction (Vetsikas & Selman, 2003), but initial predictions were essentially points.

The majority of averaging agents fixed a pool of prior games, and did not update the averages during the finals. An exception was cuhk, which employed a moving average of the previous ten games in the current round, or from previous rounds at the beginning of a new round.

UMBCTAC reported employing mean prices as predictions with respect to decisions about trips of two or more days, but median prices (which tended to be lower) for decisions about one-day trips. For the semifinals they based their statistics on the last 100 seeding games. For the finals their dataset comprised the 14 games of their semifinal heat. In our analysis below, we attribute predictions to UMBCTAC based on the mean values from these samples.

The approach taken by SouthamptonTAC (He & Jennings, 2003) was unique among TAC agents. The SouthamptonTAC designers partitioned the seeding-round games into three categories, "competitive", "non-competitive", and "semi-competitive". They then specified a reference price for each type and day of hotel in each game category. The agent then chooses a category for any game based on its monitoring of recent game history. In the actual tournament, Southampton-TAC began the semifinals predicting the semi-competitive reference prices, maintaining this stance until switching to non-competitive for the last eight games of the finals.

## 6.2 Machine Learning

A couple of TAC-02 agents employed machine learning techniques to derive relationships between observable parameters and resulting hotel prices. The premise of this approach is that game-specific features provide potentially predictive information, enabling the agent to anticipate hotel price directions before they are manifest in price quotes themselves. Not surprisingly, as noted in Section 5.2, the two learning agents employed more kinds of information than typical TAC-02 agents.

ATTac predicts prices using a sophisticated boosting-based algorithm for conditional density estimation (Stone et al., 2003). Development of the technique was expressly motivated by the TAC price-prediction problem, though the resulting algorithm is quite general. ATTac learns a predictor for each hotel type and day category (i.e., days 1 and 4 are treated symmetrically, as are 2 and 3). The predictor applied at the beginning of the game maps the following features into a predicted price for that hotel:

- identity of agents participating in the game,

- initial flight prices, and

- closing time of each hotel room.

Since the hotel closing times are unknown at game start, this predictor induces a distribution over price predictions, based on the distribution of hotel closing sequences. This distribution constitutes ATTac's initial price prediction.

kavayaH (Putchala et al., 2002) predicts initial hotel prices using neural networks trained via backpropagation. The agent has a separate network for each hotel. The output of each network is one of a discrete set of prices, where the choice set for each hotel (type, day) was specified by kavayaH's designers based on historical prices. The inputs for each network are based on initial flight prices, specifically thresholded differences between flights on adjacent days. For example, hotel T1 might have a binary input that indicates whether the price difference between inflights on days 1 and 2 is greater than 50. Hotel S2 might have this input, as well as another based on the difference in flight prices on days 2 and 3. kavayaH's designers selected the most relevant inputs based on experiments with the agent.

---

4. Several of these agents complement this simple initial prediction with a relatively sophisticated approach to interim prediction, using the evidence from price quotes to gradually override the initial estimate. All else equal, of course, straightforwardness is an advantage. Indeed, simplicity was likely a significant ingredient of livingagents's success in TAC-01 (Fritschi & Dorer, 2002; Wellman et al., 2003).

### 6.3 Competitive Analysis

Walverine's overall approach to TAC markets is to presume that they are well-approximated by a competitive economy (Cheng et al., 2004). Its method for predicting hotel prices is a literal application of this assumption. Specifically, it calculates the *Walrasian competitive equilibrium* of the TAC economy, defined as the set of prices at which all markets would clear, assuming other agents behave as price takers (Katzner, 1989). Taking into account the exogenously determined flight prices, Walverine finds a set of hotel prices that support such an equilibrium, and returns these values as its prediction for the hotels' final prices.

Let $\boldsymbol{p}$ be a vector of hotel prices, consisting of elements $p_{h,i}$ denoting the price of hotel type $h$ on day $i$. Let $x_{h,i}^j(\boldsymbol{p})$ denote agent $j$'s demand for hotel $h$ day $i$ at these prices, and write the vector of such demands as $\boldsymbol{x}^j(\boldsymbol{p})$. *Aggregate demand* is simply the sum of agent demands, $\boldsymbol{x}(\boldsymbol{p}) = \sum_j \boldsymbol{x}^j(\boldsymbol{p})$.

Prices $\boldsymbol{p}$ constitute a *competitive equilibrium* if aggregate demand equals aggregate supply for all hotels. Since there are 16 rooms available for each hotel on each day, we have that in competitive equilibrium, $\boldsymbol{x}(\boldsymbol{p}) = \boldsymbol{16}$.

Starting from an initial guess $\boldsymbol{p}^0$, Walverine searches for equilibrium prices using the *tatonnement* protocol, an iterative price adjustment mechanism originally conceived by Walras (Arrow & Hahn, 1971). Given a specification of aggregate demand, tatonnement iteratively computes a revised price vector according to the following difference equation:

$$\boldsymbol{p}^{t+1} = \boldsymbol{p}^t + \alpha^t[\boldsymbol{x}(\boldsymbol{p}^t) - \boldsymbol{16}], \qquad (2)$$

Although equilibrium prices are not guaranteed to exist given the discreteness and complementarities of the TAC environment, we have found that this procedure typically produces an approximate equilibrium well within the 300 iterations Walverine devotes to the prediction calculation.

A critical step of competitive analysis is determining the aggregate demand function. Walverine estimates $\boldsymbol{x}(\boldsymbol{p})$ as the sum of (1) its own demand based on the eight clients it knows about, and (2) the expected demand for the other agents (56 clients), based on the specified TAC distribution of client preferences. Our calculation of expected demand for the others is exact, modulo a summarization of entertainment effects, given an assumption that agent demands are separable by client (Cheng et al., 2004). This assumption is true at the beginning of the game (hence for initial prediction), but is invalidated once agents accumulate holdings of flights and hotels. Although the analytic expression of this expected demand is somewhat complicated, deriving it is not conceptually or computationally difficult.

Note that the larger component of Walverine's demand estimation is an expectation over the defined distribution of client preferences. Therefore, the prices it derives should properly be viewed as an equilibrium over the expectation, rather than the expected equilibrium prices. The latter might actually be a more appropriate measure for price prediction. However, since expected equilibrium is much more computationally expensive than equilibrium of the expectation (and we suspect the difference would be relatively small for 56 i.i.d. clients), we employ the simpler measure.

## 7. Predictions

As part of the survey, entrants provided the predictions their agents actually employed in the TAC-02 finals and semifinals: a total of 60 games. In many cases, predictions are constant (i.e., the same for every game), so it is straightforward to evaluate them with respect to the full slate of final and semifinal games. In two of the cases where initial predictions change every game (ATTac and Walverine), entrants were able to construct what their agent *would have predicted* for each of these games, whether or not they actually participated. In one case (kavayaH), we have partial data. kavayaH reported its predictions for the 32 final games, and for the semifinal heat in which it participated (H1), except for one game in which its predictor crashed.

We include two versions of ATTac, corresponding to predictors learned from the 2001 and 2002 preliminary rounds. ATTac01 and ATTac02, respectively, represent the prediction functions employed in the TAC-01 and TAC-02 finals. In applying the ATTac01 predictor to the TAC-02 finals, its use of agent identity information was disabled.

The price vectors supplied by entrants and employed in our analysis are presented in Table 3. Prices are rounded to the nearest integer for display, though our analysis employed whatever precision was provided. Agents who condition on game-specific information produce distinct vectors in each instance, so are not tabulated here.

| Agent | S1 | S2 | S3 | S4 | T1 | T2 | T3 | T4 |
|-------|----|----|----|----|----|----|----|----|
| harami | 21 | 58 | 80 | 16 | 47 | 108 | 101 | 64 |
| livingagents | 27 | 118 | 124 | 41 | 73 | 163 | 164 | 105 |
| PackaTAC | 21 | 116 | 119 | 38 | 76 | 167 | 164 | 97 |
| RoxyBot[5] | 20 | 103 | 103 | 20 | 76 | 152 | 152 | 76 |
| sics | 30 | 100 | 100 | 40 | 95 | 160 | 155 | 110 |
| WhiteBear | 19 | 102 | 96 | 28 | 75 | 144 | 141 | 81 |
| SouthamptonTAC "S" | 50 | 100 | 100 | 50 | 100 | 150 | 150 | 100 |
| SouthamptonTAC "N" | 20 | 30 | 30 | 20 | 50 | 80 | 80 | 50 |
| UMBCTAC semifinals | 20 | 133 | 124 | 45 | 83 | 192 | 158 | 110 |
| UMBCTAC finals | 37 | 75 | 87 | 29 | 113 | 141 | 95 | 71 |
| Actual Mean | 68 | 85 | 97 | 52 | 121 | 124 | 154 | 109 |
| Actual Median | 9 | 48 | 38 | 8 | 59 | 105 | 98 | 59 |
| Best Euc Dist | 18 | 73 | 57 | 15 | 71 | 111 | 95 | 69 |
| Best EVPP | 28 | 51 | 67 | 0 | 80 | 103 | 100 | 84 |
| Walverine const | 28 | 76 | 76 | 28 | 73 | 113 | 113 | 73 |

Table 3: Predicted price vectors: Shoreline Shanties, followed by Tampa Towers, each for days 1–4. The first ten rows represent predictions employed by agents in the tournament. The last five represent various benchmarks, discussed below.

The first six rows of Table 3 (harami through WhiteBear) correspond to constant predictions for their associated agents. As noted above, SouthamptonTAC switched between two prediction vectors: "S" represents the reference prices for its "semi-competitive" environment, and "N" its "non-competitive" prices. UMBCTAC as well switched prediction vectors within the 60 games—in their case introducing for the finals a prediction based on average semifinal (H1) prices.

The rows labeled "Actual Mean" and "Actual Median", respectively, present the average and median hotel prices actually resulting from the 60 games of interest. Although clairvoyance is obviously not an admissible approach to prediction, we include them here as a benchmark. In a direct sense, the actual central tendencies represent the best that agents taking the historical averaging approach can hope to capture.

The price vectors labeled "Best Euc Dist", "Best EVPP", and "Walverine const" are discussed in Section 8.2.

# 8. Evaluating Prediction Quality

## 8.1 Accuracy Measures

It remains now to assess the efficacy of the various prediction approaches, in terms of the agents' price predictions in the actual TAC-02 final and semifinal games. In order to do so, we require some measure characterizing the accuracy of a prediction $\hat{p}$ given the actual prices $p$ of a given game.

### 8.1.1 EUCLIDEAN DISTANCE

A natural measure of the closeness of two price vectors is their Euclidean distance:

$$d(\hat{\boldsymbol{p}}, \boldsymbol{p}) \equiv \left[ \sum_{(h,i)} (\hat{p}_{h,i} - p_{h,i})^2 \right]^{1/2},$$

where $(h, i)$ indexes the price of hotel $h \in \{\mathsf{S}, \mathsf{T}\}$ on day $i \in \{1, 2, 3, 4\}$. Clearly, lower values of $d$ are preferred, and for any $\boldsymbol{p}$, $d(\boldsymbol{p}, \boldsymbol{p}) = 0$.

Calculating $d$ is straightforward, and we have done so for all of the reported predictions for all 60 games. Note that if $\hat{\boldsymbol{p}}$ is in the form of a distribution, the Euclidean distance of the mean provides a lower bound on the average distance of the components of this distribution. Thus, at least according to this measure, our evaluation of distribution predictions in terms of their means provides a bias in their favor.

It is likewise the case that among all constant predictions, the actual mean $\bar{\boldsymbol{p}}$ for a set of games minimizes the aggregate *squared* distance for those games. That is, if $\boldsymbol{p}^j$ is the actual price vector for game $j$, $1 \le j \le N$,

$$\bar{\boldsymbol{p}} \equiv \frac{1}{N} \sum_{j=1}^{N} \boldsymbol{p}^j = \arg\min_{\hat{\boldsymbol{p}}} \sum_{j=1}^{N} [d(\hat{\boldsymbol{p}}, \boldsymbol{p}^j)]^2.$$

---

5. RoxyBot's prediction is based on statistics from the seeding rounds, expressed as cumulative price distributions for each hotel, discretized into deciles. RoxyBot reportedly based its decisions on samples from this distribution, taking each decile value to occur with probability 0.1. This tends to overestimate prices, however, as the decile values correspond to upper limits of their respective ranges. The prediction vector presented in Table 3 (and analyzed below) corresponds to an adjusted value, obtained by dropping the top decile and averaging the remaining nine.

There is no closed form for the prediction minimizing aggregate $d$, but one can derive it numerically for a given set of games (Bose et al., 2002).

### 8.1.2 EXPECTED VALUE OF PERFECT PREDICTION

Euclidean distance $d$ appears to be a reasonable measure of accuracy in an absolute sense. However, the purpose of prediction is not accuracy for its own sake, but rather to support decisions based on these predictions. Thus, we seek a measure that relates in principle to expected TAC performance. By analogy with standard value-of-information measures, we introduce the concept of *value of perfect prediction* (*VPP*).

Suppose an agent could anticipate perfectly the eventual closing price of all hotels. Then, among other things, the agent would be able to purchase all flights immediately with confidence that it had selected optimal trips for all clients.[6] Since many agents apparently commit to trips at the beginning of the game anyway, perfect prediction would translate directly to improved quality of these choices.[7] We take this as the primary worth of predictions, and measure quality of a prediction in terms of how it supports trip choice in comparison with perfect anticipation. The idea is that VPP will be particularly high for agents that otherwise have a poor estimate of prices. If we are already predicting well, then the value of obtaining a perfect prediction will be relatively small. This corresponds to the use of standard value-of-information concepts for measuring uncertainty: for an agent with perfect knowledge, the value of additional information is nil.

Specifically, consider a client $c$ with preferences $(pa, pd, hp)$. A trip's *surplus* for client $c$ at prices $\boldsymbol{p}$, $\sigma_c(r, \boldsymbol{p})$ is defined as value minus cost,

$$\sigma_c(r, \boldsymbol{p}) \equiv v_c(r) - cost(r, \boldsymbol{p}),$$

where $cost(r, \boldsymbol{p})$ is simply the total price of flights and hotel rooms included in trip $r$. Let

$$r_c^*(\boldsymbol{p}) \equiv \arg \max_r \sigma_c(r, \boldsymbol{p})$$

denote the trip that maximizes surplus for $c$ with respect to prices $\boldsymbol{p}$. The expression

$$\sigma_c(r_c^*(\hat{\boldsymbol{p}}), \boldsymbol{p})$$

then represents the surplus of the trip *chosen based on* prices $\hat{\boldsymbol{p}}$, but *evaluated with respect to* prices $\boldsymbol{p}$. From this we can define value of perfect prediction,

$$VPP_c(\hat{\boldsymbol{p}}, \boldsymbol{p}) \equiv \sigma_c(r_c^*(\boldsymbol{p}), \boldsymbol{p}) - \sigma_c(r_c^*(\hat{\boldsymbol{p}}), \boldsymbol{p}). \tag{3}$$

Note that our *VPP* definition (3) is relative to client preferences, whereas we seek a measure applicable to a pair of price vectors outside the context of a particular client. To this end we define the *expected value of perfect prediction*, *EVPP*, as the expectation of *VPP* with respect to TAC's distribution of client preferences:

$$
\begin{aligned}
EVPP(\hat{\boldsymbol{p}}, \boldsymbol{p}) &\equiv E_c[VPP_c(\hat{\boldsymbol{p}}, \boldsymbol{p})] \\
&= E_c[\sigma_c(r_c^*(\boldsymbol{p}), \boldsymbol{p})] - E_c[\sigma_c(r_c^*(\hat{\boldsymbol{p}}), \boldsymbol{p})].
\end{aligned}
\tag{4}
$$

---

6. Modulo some residual uncertainty regarding availability of entertainment tickets, which we ignore in this analysis.

7. We compiled statistics on the temporal profile of flight purchases for the eight agents in the TAC-02 finals. Four of the agents purchased 16 flights (enough for round trips for all clients) within 45 seconds on average. All eight agents purchased more than half their flights by that time, on average. Vetsikas and Selman (2003) verified experimentally that predicting prices benefits agents who commit to flights early to a greater extent than it does those who delay flight purchases.

Note that as for $d$, lower values of *EVPP* are preferred, and for any $\boldsymbol{p}$, $EVPP(\boldsymbol{p}, \boldsymbol{p}) = 0$.

From (4) we see that computing *EVPP* reduces to computing $E_c[\sigma_c(r_c^*(\hat{\boldsymbol{p}}), \boldsymbol{p})]$. We can derive this latter value as follows. For each $(pa, pd)$ pair, determine the best trip for hotel S and the best trip for hotel T, respectively, at prices $\hat{\boldsymbol{p}}$ ignoring any contribution from hotel premium, $hp$. From this we can determine the threshold value of $hp$ (if any) at which the agent would switch from S to T. We then use that boundary to split the integration of surplus (based on prices $\boldsymbol{p}$) for these trips, with respect to the underlying distribution of $hp$. Note that this procedure is analogous to Walverine's method for calculating expected client demand (Cheng et al., 2004) in its competitive equilibrium computation.

## 8.2 Results

Figure 1 plots the thirteen agents for whom we have prediction data according to our two quality measures. With one exception, the $d$ and *EVPP* values shown represent averages over the 60 games of TAC-02 finals and semifinals. Since kavayaH predicted only 45 games, we normalized its average $d$ and *EVPP* values to account for the relative difficulty of the games it omitted compared to the games it predicted. The normalization multiplied its raw average by the ratio of prediction qualities for these game sets by another representative agent (ATTac01, which was the most favorable choice for kavayaH, though other normalizations would have produced similar results).

The two dashed lines in Figure 1 represent best-achievable constant predictions with respect to the two accuracy measures. "Best Euc Dist" minimizes average Euclidean distance, as indicated by the vertical line. For *EVPP*, we performed hill-climbing search from a few promising candidate vectors to derive a local minimum on that measure, represented by the horizontal line. Both reference vectors are provided in Table 3. Note that in principle, only agents that varied their predictions across game instances (ATTac, kavayaH, cuhk, and Walverine, and to a coarser degree, SouthamptonTAC and UMBCTAC) have the potential to perform outside the upper-right quadrant.

To assess the significance of the accuracy rankings among agents, we performed paired-sample t-tests on all pairs of agents for both of our measures. The differences between Walverine and ATTac01 do not reach a threshold of statistical significance on either measure. Walverine beats ATTac01 for $d$ at $p = .16$ while ATTac01 beats Walverine for *EVPP* at $p = .18$. Walverine significantly ($p \le .03$) outperforms all other agents on both measures. ATTac01 significantly ($p \le .01$) outperforms all other agents for *EVPP*, but for $d$ it is not statistically distinguishable ($p \ge .08$) from kavayaH, harami, or cuhk. For *EVPP*, Walverine and ATTac01 are the only agents that beat "Best *EVPP*" ($p = .015$ and $p = .048$), and "Best *EVPP*" in turn beats all other agents (all but cuhk significantly). For $d$, Walverine is the only agent to significantly ($p < .001$) beat Best Euclidean Distance, which in turn beats every other agent but ATTac01 and kavayaH. No agent but Walverine does significantly better than Actual Mean, with ATTac01, kavayaH, and harami statistically indistinguishable.

The large discrepancy in performance between ATTac01 and ATTac02 is unexpected, given that their predictors are generated from the same boosting-based learning algorithm (Stone et al., 2003). This might be explained if the 2002 preliminary rounds were somehow less predictive of the TAC-02 finals than was the case in 2001. The relative success of another learning agent, kavayaH, is evidence against this, however. The more likely hypothesis is that the 2002 agent suffered from a bug emerging from a last-minute change in computing environments.

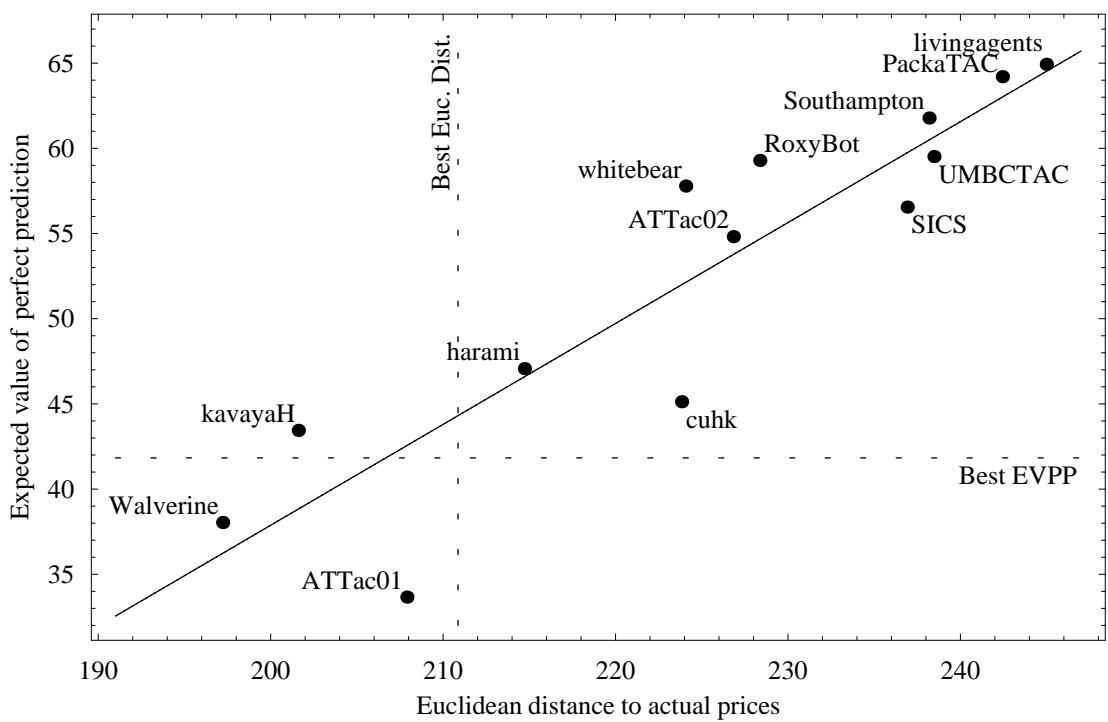

Figure 1: Prediction quality for eleven TAC-02 agents. Dashed lines delimit the accuracy achievable with constant predictions: "best Euclidean distance" and "best EVPP" for the two respective measures. The diagonal line is a least-squares fit to the points. Observe that the origin of this graph is at (190,32).

To directly evaluate a prediction in the form of pricelines, we would need to know the initial demand of this agent corresponding to the priceline. We did obtain such information from sics, but found that the accuracy of the priceline prediction according to these measures was far worse than that of the baseline prediction. Our impression is that the pricelines may well be advantageous with respect to the decisions the agents based on them, but do not improve basic accuracy. Note that *EVPP* inherently is based on an interpretation of prices as linear, thus it may not provide a proper evaluation of priceline predictions.

### 8.3 The Influence of Flight Prices

Observe that the three best price predictors—ATTac01, Walverine, and kavayaH—are exactly those agents that take flight prices into account. Initial flight prices potentially affect hotel prices through their influence on agents' early trip choices. In theory, lower flight prices should increase the tendency of agents to travel on those days, all else equal, thus increasing the prices of hotels on the corresponding days of stay. Indeed, capturing this effect of flight prices was one of the main motivations for Walverine's price-equilibrium approach. ATTac and kavayaH attempt to induce the relationship from game data. kavayaH's designers, in particular, explored neural network models based on their hypotheses about which flights were likely to affect which hotel prices (Putchala et al., 2002).

To isolate and quantify the effect of flight prices, we investigated the contribution of different factors employed by Walverine in its predictions. We defined three additional versions of Walverine's prediction model, each of which ignores some information that Walverine takes into account:

- Walv-no-cdata ignores its own client knowledge, effectively treating own demand as based on the same underlying client preference distribution assumed for the other agents.

- Walv-constF ignores the initial flight prices, assuming that they are set at the mean of the initial flight distribution (i.e., 325) in every game instance.

- Walverine const ignores its own client knowledge *and* takes flight prices at their mean rather than actual values. The result is a constant prediction vector, presented in Table 3.

Figure 2 plots the prediction qualities of these agents. Ignoring client knowledge degraded prediction quality only slightly, increasing *EVPP* from 38.0 to 38.6. Neglecting initial flight prices, however, significantly hurt predictions, increasing *EVPP* to 47.9. Ignoring both, Walverine const incurred an average *EVPP* of 49.1.

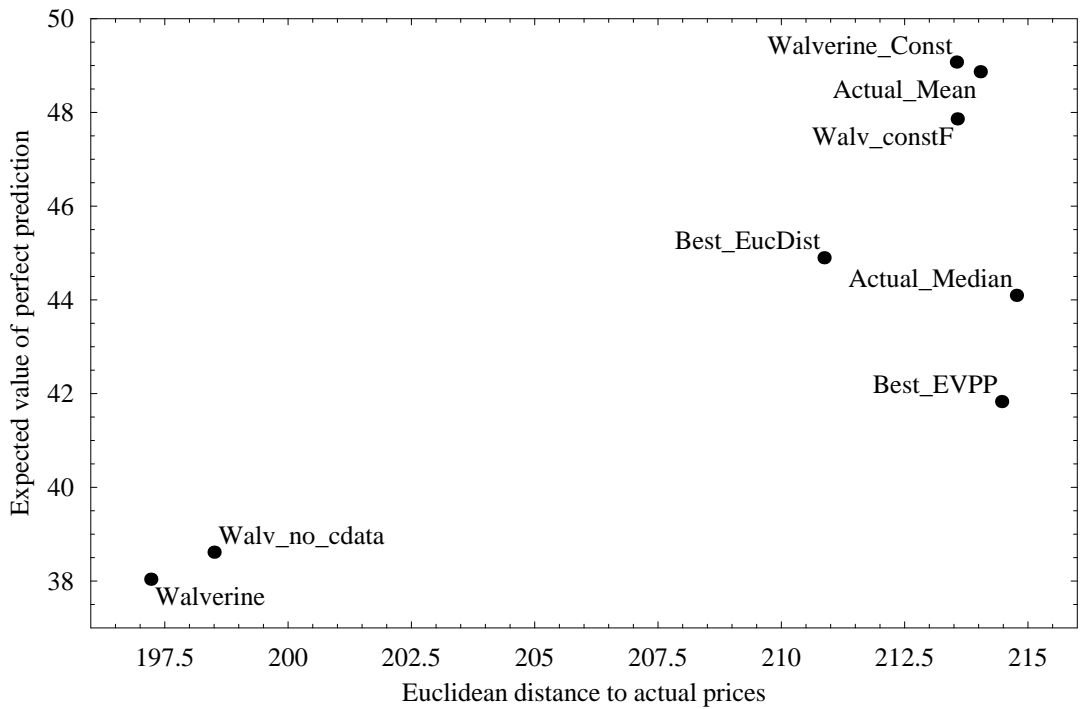

Figure 2: Prediction quality for Walverine variants and various central tendency benchmarks.

The results confirm the predictive value of flight prices. On the *EVPP* measure, Walverine does not gain a significant advantage from considering own client data, but cannot beat "Best *EVPP*" without considering initial flight prices. For $d$, client data does make a significant ($p = .03$) difference when also considering flight data. Flight data significantly ($p < .001$) affects **Walverine**'s prediction quality for the $d$ metric, regardless of client data.

### 8.4 Relating Accuracy to Performance

As indicated by the scatter plot of Figure 1, our two accuracy measures are highly correlated ($\rho = 0.908$). Given that *EVPP* is value-based, this does suggest that accuracy translates somewhat proportionally to performance. However, *EVPP* is a highly idealized proxy for actual scores, and so does not definitively establish the relation of prediction accuracy to overall TAC performance.

Such a relation was observed in prior work by Stone et al. (2003), who evaluated the predictive accuracy and overall performance of four variations on **ATTac01** in an experimental trial. Employing a different measure of prediction quality than ours above, they found that average score was related monotonically to average predictive accuracy.

In an effort to more directly connect our accuracy measures to the bottom line, we regressed the actual TAC-02 game scores against the accuracy measures for all reported predictions—one data point per agent per game. We controlled for favorability of random client preferences, employing the same summary statistics used to construct the "client preference adjustment" in our analysis of the TAC-01 tournament (Wellman et al., 2003). In two separate regressions, we found highly significant coefficients ($p < 10^{-10}$) for both $d$ and *EVPP*. Predictive accuracy explained score variance quite poorly ($R^2 \leq 0.12$), however, as might be expected given all the other unmodeled variation across agents.

To reduce the variation, we undertook a series of controlled trials involving variants of **Walverine** (Cheng et al., 2004). Each trial comprised a set of games with a fixed configuration of agents. The agents were constant within trials, but varied across, as they were conducted weeks or months apart while **Walverine** was undergoing modifications. For each trial, we regressed the actual score of the first agent on *EVPP*, controlling as above for favorability of random client preferences. We considered only one agent per game, since the data points for the other agents would be dependent given their common game instance. The results of our linear regression are summarized in Table 4.

| Trial | $N$ | Mean *EVPP* | *EVPP* Coeff | $R^2$ |
|-------|-----|-------------|--------------|-------|
| 1 | 200 | 70.4 | -8.89 | 0.57 |
| 2 | 151 | 32.2 | -11.59 | 0.26 |
| 3 | 110 | 59.5 | -10.26 | 0.65 |

Table 4: Regression of score on *EVPP* in three trials.

The *EVPP* coefficient was highly significant in all cases ($p < 10^{-5}$). Note that since *EVPP* is measured per-client in the same units as scores, a direct translation would equate reducing *EVPP* by one with an increase of eight score points. Our regressions yielded coefficients ranging from -8.89 to -11.59, which we take as a rough confirmation of the expected relationship. If anything, the results indicate that *EVPP* understates the value of prediction—which we might expect since it addresses only initial trip choice. Interestingly, the regression model seems to provide a better fit (as measured by $R^2$) for the trials involving worse price predictors (as measured by mean *EVPP*). This suggests that as prediction is optimized, other unmodeled factors may have relatively greater incremental influence on score.

It should be noted that these games appear to be quite unrepresentative of TAC tournament games. Since the agents are all versions of Walverine, they tend to make trip choices on the same basis.

## 9. Limitations

It is important to emphasize several limitations of this study, which must qualify any conclusions drawn about the efficacy of prediction methods evaluated here.

First, we have focused exclusively on initial price prediction, whereas many agents placed greater emphasis on the interim prediction task.

Second, in many cases we have represented agents' predictions by an abstraction of the actual object produced by their prediction modules. In particular, we reduce probability distributions to their means, and consider only the first unit of a priceline prediction. More generally, we do not account for the very different ways that agents *apply* the predictions they generate. Without question even the measure we introduce, *EVPP*, is inspired by our thinking in terms of how Walverine uses its predictions. Perhaps measures tailored to the processes of other agents would (justifiably) show their predictions in a more favorable light.

Third, it should be recognized that despite the desirability of isolating focused components of an agent for analysis, complete separation is not possible in principle. Prediction evaluation is relative not only to how the agent uses its prediction, but also how it makes other tradeoffs (e.g., when it commits to flights), and ultimately its entire strategy, in all its complexity. Studies such as this must strive to balance the benefits of decomposition with appreciation for the interconnections and synergies among elements of a sophisticated agent's behavior.

## 10. Conclusions

We have presented a comprehensive survey of approaches to initial price prediction in TAC-02, with quantitative analysis of their relative accuracy. Our analysis introduces a new measure, expected value of perfect prediction, which captures in an important sense the instrumental value of accurate predictions, beyond their nominal correspondence to realized outcomes.

We draw several conclusions about price prediction in TAC from this exercise. First, the results clearly demonstrate that instance-specific information can provide significant leverage over pure background history. Three agents use instance-specific information to perform better on at least one measure than any constant prediction. In particular, the initial flight prices provide substantial predictive value. This can be induced and verified empirically, as seen through the success of the machine-learning agents. The predictive value flows from the influence of flight prices on demand for hotels, as indicated by the success of competitive analysis in capturing this relationship.

We believe it striking that a purely analytical approach, without any empirical tuning, could achieve accuracy comparable to the best available machine-learning method. Moreover, many would surely have been skeptical that straight competitive analysis could prove so successful, given the manifest unreality of its assumptions as applied to TAC. Our analysis certainly does not show that competitive equilibrium is the best possible model for price formation in TAC, but it does demonstrate that deriving the shape of a market from an idealized economic theory can be surprisingly effective.

There are several advantages to model-based reasoning, most obviously the ability to perform with minimal or no empirical data. Even when historical information is available, it can be misleading to rely on it in a nonstationary environment. A tournament setup like TAC naturally violates stationarity, as the agent pool evolves over time, through selection as well as individual learning and development. Of course, dealing with time-variance, particularly in multiagent environments, is an active area of current research, and ultimately the best methods will combine elements of model-based and data-based reasoning.

Finally, we suggest that the present study represents evidence that research competitions can in the right circumstances produce knowledge and insights beyond what might emerge from independent research efforts. We hope that additional researchers will be inspired to bring their innovative ideas on trading strategy to the next TAC, and look forward to investigating the results of such interplay in future work.

## Acknowledgments

This study would not have been possible without the generous cooperation of the TAC-02 entrants. The research was supported in part by NSF grant IIS-9988715, as well as a STIET fellowship to the third author, under an NSF IGERT grant.

## References

Arrow, K. J., & Hahn, F. H. (1971). *General Competitive Analysis*. Holden-Day, San Francisco.

Bose, P., Maheshwari, A., & Morin, P. (2002). Fast approximations for sums of distances, clustering and the Fermat-Weber problem. *Computational Geometry: Theory and Applications*, *24*, 135–146.

Boyan, J., & Greenwald, A. (2001). Bid determination in simultaneous auctions: An agent architecture. In *Third ACM Conference on Electronic Commerce*, pp. 210–212, Tampa, FL.

Cheng, S.-F., Leung, E., Lochner, K. M., O'Malley, K., Reeves, D. M., & Wellman, M. P. (2004). Walverine: A Walrasian trading agent. *Decision Support Systems*, *To appear*.

Fritschi, C., & Dorer, K. (2002). Agent-oriented software engineering for successful TAC participation. In *First International Joint Conference on Autonomous Agents and Multi-Agent Systems*, Bologna.

Greenwald, A. (2003a). The 2002 trading agent competition: An overview of agent strategies. *AI Magazine*, *24*(1), 83–91.

Greenwald, A. (2003b). Bidding under uncertainty in simultaneous auctions. In *IJCAI-03 Workshop on Trading Agent Design and Analysis*, Acapulco.

He, M., & Jennings, N. R. (2003). SouthamptonTAC: An adaptive autonomous trading agent. *ACM Transactions on Internet Technology*, *3*, 218–235.

Katzner, D. W. (1989). *The Walrasian Vision of the Microeconomy*. University of Michigan Press.

Putchala, R. P., Morris, V. N., Kazhanchi, R., Raman, L., & Shekhar, S. (2002). kavayaH: A trading agent developed for TAC-02. Tech. rep., Oracle India.

Sadeh, N., Arunachalam, R., Eriksson, J., Finne, N., & Janson, S. (2003). TAC-03: A supply-chain trading competition. *AI Magazine*, *24*(1), 92–94.

Stone, P. (2002). Multiagent competitions and research: Lessons from RoboCup and TAC. In *Sixth RoboCup International Symposium*, Fukuoka, Japan.

Stone, P., & Greenwald, A. (2004). The first international trading agent competition: Autonomous bidding agents. *Electronic Commerce Research*, *To appear*.

Stone, P., Schapire, R. E., Littman, M. L., Csirik, J. A., & McAllester, D. (2003). Decision-theoretic bidding based on learned density models in simultaneous, interacting auctions. *Journal of Artificial Intelligence Research*, *19*, 209–242.

Vetsikas, I. A., & Selman, B. (2003). A principled study of the design tradeoffs for autonomous trading agents. In *Second International Joint Conference on Autonomous Agents and Multi-Agent Systems*, pp. 473–480, Melbourne.

Wellman, M. P., Greenwald, A., Stone, P., & Wurman, P. R. (2003). The 2001 trading agent competition. *Electronic Markets*, *13*, 4–12.